\documentclass[lettersize,journal]{IEEEtran}
\usepackage{amsmath,amsfonts}
\usepackage{array}
\usepackage{subcaption}
\usepackage{stfloats}
\usepackage{url}
\usepackage{verbatim}
\usepackage{amsmath, amssymb, xspace,  comment}
\usepackage{tikz, xcolor}
\usepackage{algpseudocode}
\usepackage[linesnumbered]{algorithm2e}
\usepackage{colortbl}
\usepackage{textcase} 
\usepackage{enumitem}
\usepackage{balance}
\usepackage{graphicx}
\usepackage{multirow}
\usepackage{hhline}
\usepackage{cite}
\usepackage{makecell}
\usepackage[hidelinks]{hyperref}
\usepackage{booktabs}
\usepackage{orcidlink}

\def\BibTeX{{\rm B\kern-.05em{\sc i\kern-.025em b}\kern-.08em
    T\kern-.1667em\lower.7ex\hbox{E}\kern-.125emX}}

\newcommand*\cib[1]{\tikz[baseline=(char.base)]{
                            \node[shape=circle,fill=black,text=white,draw,inner sep=0.3pt] (char) {#1};}}

\newcommand{\eg}{{\em e.g.,}\xspace}
\newcommand{\ie}{{\em i.e.,}\xspace}

\newcommand{\BfPara}[1]{{\vspace{0.5ex}\noindent\bf#1.}\xspace}

\newcommand{\ours}{{\NoCaseChange{{QuScore}}}\xspace}

\usepackage{tcolorbox}
\tcbuselibrary{theorems}

\newtcbtheorem[]{observation}{\textbf{Observation}}%
{colframe=white!95!black,fonttitle=\bfseries,coltitle=black, boxsep = 2pt, left = 0pt, right = 0pt, top = 1pt, bottom = 2pt, boxrule = 0pt, bottomrule = 0pt, toprule = 0pt}{th}{}

\usepackage{amsthm}
\usepackage{color}
\newtheoremstyle{styledef}%
{\topsep}
{\topsep}
{}
{}
{\bfseries}
{.}
{3pt}
{}
\theoremstyle{styledef}

\newcommand{\observationdef}[2][0.47\textwidth]{
  \vspace{2ex}\par\noindent\tikzstyle{mybox} = [fill=gray!10,
   thick,rectangle,inner sep=4pt,path picture={\fill [white!40!black] ([xshift=-4.25cm]path picture bounding box.north) rectangle (path picture bounding box.south west);}]
  \begin{tikzpicture}
   \node [mybox] (box){%
    \begin{minipage}{#1}{ #2}\end{minipage}
   };
  \end{tikzpicture}
}

\usepackage{soul}
\newcommand{\mo}[2]{{\leavevmode\color{red}{{\bf Q: #1 \xspace}}}{\hl{#2}}}
\newcommand{\eldor}[1]{\textcolor{blue}{#1}}

\hypersetup{
	colorlinks=true,
	urlcolor=blue!70!black,
	linkcolor=purple,
	citecolor=blue!80!black,
}

\begin{document}
\pagestyle{empty}

\title{Microbial Genetic Algorithm-based Black-box Attack against Interpretable Deep Learning Systems}

\author{
    \IEEEauthorblockN{Eldor Abdukhamidov\IEEEauthorrefmark{1}, 
    Mohammed Abuhamad\IEEEauthorrefmark{2}, 
    Simon~S.~Woo \IEEEauthorrefmark{1}, 
    Eric Chan-Tin\IEEEauthorrefmark{2}, 
    Tamer~Abuhmed\IEEEauthorrefmark{1}
   }\\
    \IEEEauthorblockA{\IEEEauthorrefmark{1}Sungkyunkwan University University}
    \IEEEauthorblockA{\IEEEauthorrefmark{2}Loyola University Chicago}


        \IEEEcompsocitemizethanks{\IEEEcompsocthanksitem  Eldor Abdukhamidov and Tamer ABUHMED are with Department of Computer Science and Engineering, Sungkyunkwan University, Suwon, South Korea.\protect
        (E-mail: abdukhamidov@skku.edu, 
        E-mail: tamer@skku.edu). Simon S. Woo is with Department of Artificial Intelligence and Department of Applied Data Science, Sungkyunkwan University, Suwon, South Korea.\protect
        (E-mail: swoo@g.skku.edu). Mohammed Abuhamad and Eric Chan-Tin are with Department of Computer Science, Loyola University, Chicago, USA.\protect
        (E-mail: mabuhamad@luc.edu,
        E-mail: chantin@cs.luc.edu).\\     }
    }
       


\maketitle

\IEEEtitleabstractindextext{%
\begin{abstract}
Deep neural network (DNN) models are susceptible to adversarial samples in white and black-box environments.
Although previous studies have shown high attack success rates, coupling DNN models with interpretation models could offer a sense of security when a human expert is involved, who can identify whether a given sample is benign or malicious.
However, in white-box environments, interpretable deep learning systems (IDLSes) have been shown to be vulnerable to malicious manipulations.
In black-box settings, as access to the components of IDLSes is limited, it becomes more challenging for the adversary to fool the system.
In this work, we propose a \textbf{\ul{Qu}}ery-efficient \textbf{\ul{Score}}-based black-box attack against IDLSes, \ours{}, which requires no knowledge of the target model and its coupled interpretation model. 
\ours{} is based on transfer-based and score-based methods by employing an effective microbial genetic algorithm.
Our method is designed to reduce the number of queries necessary to carry out successful attacks, resulting in a more efficient process. By continuously refining the adversarial samples created based on feedback scores from the IDLS, our approach effectively navigates the search space to identify perturbations that can fool the system.
We evaluate the attack's effectiveness on four CNN models (Inception, ResNet, VGG, DenseNet) and two interpretation models (CAM, Grad), using both ImageNet and CIFAR datasets.
Our results show that the proposed approach is query-efficient with a high attack success rate that can reach between 95\% and 100\%  and transferability with an average success rate of 69\% in the ImageNet and CIFAR datasets.
Our attack method generates adversarial examples with attribution maps that resemble benign samples.
We have also demonstrated that our attack is resilient against various preprocessing defense techniques and can easily be transferred to different DNN models.
\end{abstract}

}

\IEEEdisplaynontitleabstractindextext
\IEEEpeerreviewmaketitle

\section{Introduction} \label{sec:intro}

The tremendous development and deployment of deep learning methods in practice have brought the attention of adversaries to exploit specific vulnerabilities within the application pipeline to compromise the results or lead the model to misbehave.
There have been numerous studies on the robustness of deep learning models and their behavior in adversarial settings, specifically adversarial examples.
Adversarial examples can be used for various adversarial purposes such as poisoning, evasion, model extraction, and inference.

Increasing the security level and understanding of the inner workings of DNN models, several studies have shown that model interpretability has significant importance in both theory and practice in terms of resilience against adversarial attacks. 
Interpretable Deep Learning Systems (IDLSes) can be examples of DNN models with interpretable knowledge representations.
Furthermore, existing attacks~\cite{madry2017towards, kurakin2016adversarial, xiao2018spatially} against DNN models are found to be ineffective as the interpretation can reveal adversarial manipulations, \ie added perturbations to the example input.

However, recent studies have shown that IDLSes are still susceptible to adversarial manipulations in white-box settings~\cite{zhang2020interpretable,abdukhamidov2021advedge,abdukhamidov2022interpretations}.
It is possible to generate adversarial examples that can mislead the target DNN model and deceive its coupled interpreter simultaneously.
For example, an adversarial sample can be misclassified by the target DNN model and interpreted identically to its benign interpretation.

Attacks~\cite{zhang2020interpretable,abdukhamidov2021advedge, abdukhamidov2022interpretations} are based on the white-box scenario, in which the attacker has complete knowledge of the target model and can achieve a high attack success rate with high confidence.
In practice and in most circumstances, the target model is unreachable.
Therefore, this form of attack, \ie white-box attack, has limited practicality.
On the other hand, black-box attacks assume that the adversary can only query the model and access the output without extended knowledge about any of the system's components or the model's parameters.
The attack is therefore more realistic in a black-box environment. 
The most common examples of this type of attack are transfer-based \cite{dong2019evading, huang2019enhancing, szegedy2013intriguing} and score-based attacks\cite{wang2020mgaattack, alzantot2019genattack}. Transfer-based methods involve the use of multiple image transformation techniques to enhance the transferability of adversarial examples. The score-based approach is model-agnostic and depends solely on the predicted scores of the model, such as class probabilities or logits. These attacks estimate the gradient numerically using the prediction of the model at a conceptual level.
In black-box settings, attacking IDLSes is still an unexplored field, with many challenges to which this work contributes.

In this paper, we conduct an in-depth investigation of the security of IDLSes in a black-box environment.
Specifically, we propose a novel black-box attack that generates adversarial examples to mislead the target DNN models and their coupled interpreters.
We use transfer-based strategies to take advantage of the transferability of adversarial samples across various DNN models. In addition, we use score-based techniques to efficiently guide the search process.
This ensures that our attack is query-efficient and practical in real-world scenarios.

By evaluating our approach against four DNN models (\ie Inception-V3, DenseNet-169, VGG-19, and ResNet-50) and two interpreters (\ie CAM and Grad) on various datasets, we show the possibility and practicality of generating successful adversarial examples with accurate interpretations in a black-box environment.
In all test cases, the attack had a success rate of more than 95\% with a high degree of similarity of more than 80\% in interpretation.

\BfPara{Contributions} We summarize our contributions as follows:
\begin{itemize}[leftmargin=1em]
    \item We propose a stealthy and query-efficient black-box attack against IDLSes.
    We empirically evaluate the effectiveness of the attack from the perspective of four DNN models (\ie Inception-V3~\cite{szegedy2016rethinking}, ResNet-50~\cite{he2016deep}, Densenet-169~\cite{huang2017densely}, and VGG-19~\cite{simonyan2014very}) and two interpretation models (CAM~\cite{zhou2016learning} and Grad~\cite{simonyan2014deep}). 
    Our results show that the proposed attack has a high success rate of 95\% to 100\% in attacking target models and their interpreters on ImageNet and CIFAR datasets. This is achieved with only an average of 150 queries, demonstrating its efficiency.
    \item We present the resilience of the proposed attack when using four defense mechanisms: \ie bit depth compression, median smoothing, JPEG compression, and random resizing/padding. The results show that \ours achieved a success rate ranging from 81\% to 100\% and an average of 137 queries when using defenses.
    \item We evaluated the transferability of the attack against different DNN models.
    The results show high transferability across multiple datasets and various deep neural network models, indicating its effectiveness in generating adversarial examples that can fool different models with an average success rate of 69\% and an IoU score of 90\%.
\end{itemize}

\BfPara{\textbf{Organization}} Our paper is organized as follows:
\autoref{sec:related} surveys related research studies;
 \autoref{sec:methods} describes the notations and terms used in the paper and presents the proposed attack and its underlying mechanisms;  
 \autoref{sec:evaluation} provides the results of the attack effectiveness, robustness, and transferability against DNN and interpretation models; 
 \autoref{sec:discussion} explains the existing limitations and future work; and \autoref{sec:conc} concludes the paper.

\section{Related Work} \label{sec:related}
This section provides an overview of the related research on attacks against DNN models.
Specifically, it covers previous work on both white-box and black-box attacks that employ various techniques, including transfer-based attacks, interpretation-based attacks, and score-based attacks.

\BfPara{Transfer-based attacks} Previous research has shown that adversarial samples generated to attack specific DNN models can also be used to fool other DNN models.
Such attacks, known as transfer-based attacks~\cite{szegedy2013intriguing}, use adversarial inputs generated by white-box attacks against DNNs to attack other black-box models.
One study~\cite{huang2019enhancing} proposed a method that adds perturbations to the hidden layers of a model to enhance the transferability of adversarial samples.
Another research work~\cite{dong2019evading} proposed a technique to improve adversarial transferability by convolving the gradient through a specific kernel.

\BfPara{Interpretation-based attacks} Adversarial attacks based on interpretation generate samples that can deceive both the target DNN models and their interpreters simultaneously~\cite{zhang2020interpretable,abdukhamidov2022interpretations, abdukhamidov2022black,juraev2022depth,abdukhamidov2021advedge}.
A recent study~\cite{abdukhamidov2021advedge} proposed white-box attacks called AdvEdge and AdvEdge$^+$ against DNN models and their coupled interpreters.
These attacks demonstrate the vulnerability of DNN models that rely on interpretable features for decision making and highlight the need to consider the interpretability of DNN models in addition to their accuracy and robustness.
The proposed attack provides a valuable tool to assess the interpretability of DNN models and their susceptibility to adversarial attacks.

\BfPara{Score-based attacks} Heuristic methods, such as evolution strategies and genetic algorithms, have been used to devise adversarial attacks that can generate visually imperceptible samples to deceive DNN models~\cite{alzantot2019genattack}.
GenAttack is a gradient-free optimization attack that can generate adversarial samples against black-box models with fewer queries.
Another study proposed a query-efficient attack called MGAAttack~\cite{wang2020mgaattack}, which uses transfer-based techniques to improve its efficacy.
These attacks highlight the vulnerability of DNN models to adversarial attacks and the need to develop more robust defense mechanisms to enhance their security.
By studying these attacks, researchers can identify weaknesses in DNN models and devise better defenses against adversarial attacks.

\section{\NoCaseChange{\ours{}}: Methods} \label{sec:methods}
The section describes \ours{} in black-box settings with a detailed explanation of the adopted methods. 

\subsection{Concepts and Notations} \label{subsec:fundamental}
In this subsection, we introduce the notation, terms, and symbols used in the paper.  

\BfPara{Classifier} Image classification is the primary focus of this work, and we employ two types of DNN models: white-box and black-box models.
In the paper, $f(x)=y \in Y$ denotes a classifier in a black box setting (target DNN model), and $f'(x)=y \in Y$ denotes a classifier in a white box setting (source DNN model), where $y$ refers to a single category from a set of categories $Y$.

\BfPara{Interpreter} We adopt an existing interpretation model $g$ to produce an interpretation map $m$ to display the importance of features for a sample $x$, which are found by a classifier $f$: $g(x; f)= m$.
For our approach, post hoc interpretability is used~\cite{dabkowski2017real, fong2017interpretable, karpathy2015visualizing, murdoch2018beyond}, where an interpretation model $g$ obtains information about the sample input $x$ and its classification decision by the classifier $f$ to generate an attribution map.
This type of interpretation requires another model to interpret the decision process of the current classification model.

\BfPara{Adversarial Attack} Pixel perturbation attack known as PGD~\cite{madry2017towards} generates an adversarial sample $\hat{x}$ with the objective of causing the source DNN model $f'$ to misclassify $\hat{x}$ into another category: $f'(\hat{x}) \neq y$.
PGD is implemented as follows:

\begin{equation*}
\begin{split}
    \hat{x}^{(i+1)} = \prod _{\mathcal{B}_{\varepsilon}(x)}\left(\hat{x}^{(i)} - \alpha \cdot ~sign(\nabla_{\hat{x}}\ell_{adv}(f'(\hat{x}^{(i)})))\right)
\end{split}
\end{equation*}
where $\prod$ is the projection operator, $\alpha$ is the learning rate, $\ell_{adv}$ is the loss function (\ie cross entropy), $\mathcal{B}_{\varepsilon}(x)$ is the norm ball limited with the specific range $\varepsilon$, and $\hat{x}^{(i)}$ is $\hat{x}$ at iteration $i$. 

\BfPara{Threat Model} In this work, we consider a black-box setting, where the adversary has limited access to the target DNN classifier $f$ (model output) and access is not given to the interpretation model $g$.
The settings resemble a realistic scenario for the attack.

\begin{figure*}[t]
    \centering
    \includegraphics[width=0.9\linewidth]{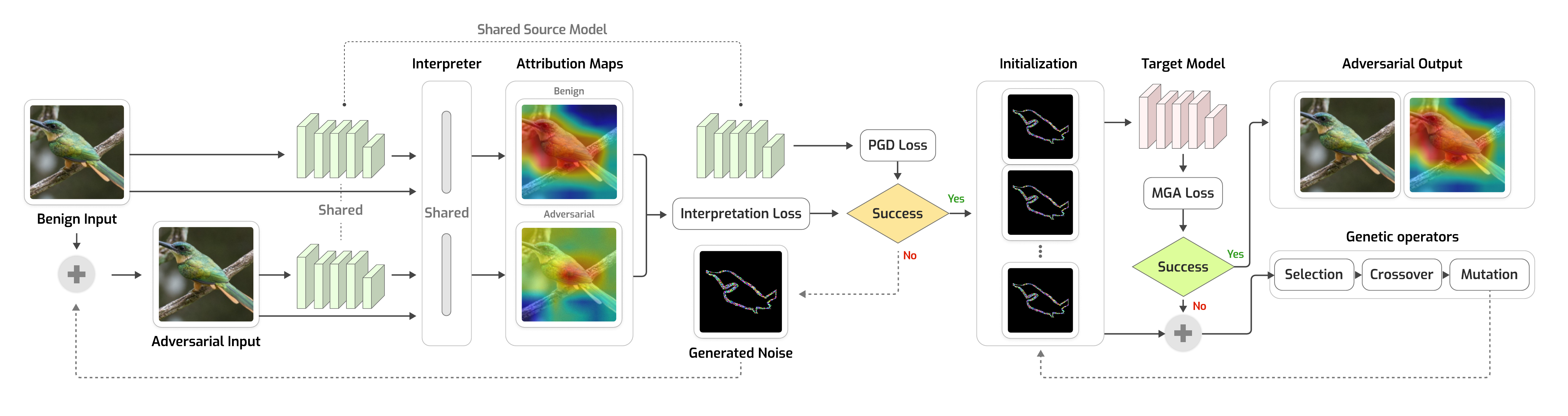}
    \caption{The framework of \ours{}: 
    After generating successful adversarial examples using the PGD attack on the source model, these examples are passed to the MGA algorithm as the initial population. Each sample in the population is passed to the target model (\ie black-box model) to check if it is effective against it; if not, two samples that have a high and lower impact on the model are selected for the crossover and mutation process to generate new samples, which are evaluated and examined against the target.
    The process is repeated until a perturbation is produced that fools the target IDLS.}
    \label{fig:blackbox_advedge}
\end{figure*}

\subsection{Attack Formulation} \label{subsec:attack_formulation}
Attacking IDLSes requires fooling both a DNN model and its associated interpretation model.
In a white-box scenario, AdvEdge~\cite{abdukhamidov2021advedge}, takes advantage of the edge information in a sample input to generate an adversarial sample to mislead a DNN classifier $f'$ and its interpretation model $g$ concurrently.
The main objective of the attack is to generate an adversarial sample $\hat{x}$ by fulfilling the following conditions: \cib{1} $\hat{x}$ should fool the source DNN model $f'$: $f'(\hat{x}) \neq y$;
\cib{2} an interpretation map $\hat{m}$ of the adversarial sample $\hat{x}$ generated by an interpreter $g$ should be similar to the interpretation map $m$ of the benign sample $x$: $g(\hat{x};f')=\hat{m}$ st. $\hat{m} \cong m$;
\cib{3} the adversarial sample $\hat{x}$ and its benign version $x$ should be visually imperceptible; \cib{4} The amount of noise is restricted to the edge of the sample image input.
At a high level, the attack framework is formulated as follows:
\begin{equation} \label{eq:nonlinear}
\begin{split}
    \min _{\hat{x}} : \Delta(\hat{x}, x) \quad s.t. \left\{ \begin{array}{lcr}
         f'(\hat{x}) \neq y, \quad st. \quad \| \hat{x} - x\|_{\infty} \in \{-\epsilon, \epsilon\}\\
         g(\hat{x}; f') = \hat{m}, \quad st. \quad \hat{m} \cong m \\
         \Delta(\hat{x}, x) \sim edge(x \cap m)
         \end{array}\right.
\end{split}
\end{equation}

The \autoref{eq:nonlinear} ensures that the prediction of the adversarial sample is not equal to the original category; an interpretation map of the adversarial sample is similar to the interpretation map of its benign version; the added perturbation lies within the edges of the sample that intersect with the interpretation map of the sample.
The attack aims to minimize the overall adversarial loss ($\ell_{adv}$) considering the classification loss
$\ell_{prd}(f'(x)) = - \log(f'(x))$ and the interpretation loss $\ell_{int}(g(x;f', m) = \|g(x;f') - m\|_{2}^2$.

The overall adversarial loss is then defined as follows:

\begin{equation*} \label{eq:mainFormula}
    \ell_{adv}= \min_{\hat{x}} \ell_{prd}(f'(\hat{x})) + \lambda ~\ell_{int}(g(\hat{x};f'), m)\text{,}
\end{equation*}
where the hyperparameter $\lambda$ balances $\ell_{prd}$ and $\ell_{int}$.
The final adversarial framework can be described as follows: 

\begin{equation} \label{eq:finalFormula}
\begin{split}
    \hat{x}^{(i+1)} = \prod _{\mathcal{B}_{\varepsilon}(x)}\left(\hat{x}^{(i)} - N_{w}~ \alpha ~ .~ sign(\nabla_{\hat{x}}\ell_{adv}(\hat{x}^{(i)}))\right)\text{,}
\end{split}
\end{equation}
where $\prod$ is the production operator, $\mathcal{B}_{\varepsilon}$ is a norm ball, $\alpha$ is the learning rate, $x$ is the sample input and $\hat{x}^{(i)}$ is the adversarial sample at the \textit{i}th iteration. $N_{w}$ is the edge operator function that is used to optimize the location and magnitude of the added perturbation:
\begin{equation*}
\begin{split}
d = \sqrt{d_{h}^2 + d_{v}^2}\\
    N_{w} = d \cap m \text{,}
\end{split}
\end{equation*}
where $d$ is an image that contains edges of the sample $x$ extracted by $d = \sqrt{d_{h}^2 + d_{v}^2}$. $d_{h}$ and $d_{v}$ contain the horizontal and vertical edge information of the sample.
The attack uses the intersection of the edges of a sample image and its interpretation map to extract important regions.

As in \autoref{eq:finalFormula}, the PGD framework~\cite{madry2017towards} is adopted to generate adversarial samples as initial seed (populations) for the MGA algorithm~\cite{harvey2009microbial} (line 1 in \autoref{alg:algorithm_mga}). These initial samples are generated using the PGD framework on the source model (not the target model $f$) and their interpreters.
Additionally, we employ the MGA algorithm to optimize adversarial samples generated against the target black-box DNN classifier $f$. The attack is summarized in \autoref{fig:blackbox_advedge}.

\subsection{\NoCaseChange{\ours{}} using Microbial Genetic Algorithm (MGA)}\label{subsec:blackbox_attack}
MGA~\cite{harvey2009microbial} is a type of genetic algorithm that populates candidate solutions based on a gradient-free optimization technique.
In the technique, a set of samples (called the population) is iteratively evolved to generate optimal candidates with higher fitness.
A sample in each iteration is referred to as a generation.
The quality of each member of the population is evaluated via a fitness function that assigns it a value based on the objective function in the optimization problem.
Samples with high fitness scores are more likely to be selected to produce the next generation through crossover and mutation.
However, the direct implementation of MGA is found to be ineffective in terms of model interpretation.
Hence, various modifications have been made to improve its efficacy.


\RestyleAlgo{ruled}
\begin{algorithm}[t]
\caption{\ours{} in Black-box 
Settings}\label{alg:algorithm_mga}
\KwData{Source DNN $f'$, interpreter $g$, input $x$, original category $y$, perturbation threshold $\epsilon$, mutation rate $mr$, crossover rate $cr$, population size $n$, generation $G$, target DNN $f$} 
\KwResult{Adversarial sample $\hat{x}$}
$x'$ = advedge\_attack($f'$, $g$, $x$, $n$) \\
$pop$ = init\_population($x$, $x'$, $\epsilon$) \\
\For{$g\leftarrow 1$ \KwTo $G$}{
$p_1$, $p_2$ = random\_select($pop$)\\
$v_1$, $v_2$ = get\_fitness($f$, $x$, $p_1$, $p_2$)\\
$loser$, $winner$ = sort\_by\_fitness($p_1$, $p_2$, $v_1$, $v_2$)\\
$child$ = crossover($cr$, $loser$, $winner$)\\
$child$ = mutation($mr$, $child$)\\
\If {$f(child) \neq y$}{
return $child$
}
$pop$ = update\_population($pop$, $child$)
}
\end{algorithm}


Our approach is based on the transfer-based technique~\cite{szegedy2013intriguing, dong2019evading}, in which adversarial samples generated using a particular DNN model can be used to fool other DNN models.
We generate adversarial samples against a DNN model in a white-box setting and use them as the initial population for MGA. MGA updates the initial population to produce new generations of those adversarial examples to deceive a black-box DNN model $f$ and mislead its coupled interpreter $g$.

Detailed information on the attack is described in \autoref{alg:algorithm_mga}.
The attack consists of genetic algorithm operators: \textit{initialization} (line 1--2), \textit{selection} (line 4--6), \textit{crossover} (line 7), \textit{mutation} (line 8), and \textit{population update} (line 12).

\textbf{\ul{Initialization.}} Seeding the initial population is the first phase of MGA. Although the process is executed only once during the attack process, the initial population is crucial to the convergence of the technique.
The population with an optimal solution helps the technique converge quickly.
In our case, we generate adversarial samples using the AdvEdge attack, which is explained in \autoref{subsec:attack_formulation} and provide them as the initial population of MGA $\Psi: ~\{\psi_1, \psi_2, \dots, \psi_m\} $, where $m$ is the size of the population.

\textbf{\ul{Fitness function.}} This is also known as the evaluation phase.
It is used to assess the quality of individuals in a population and help evolve towards the optimal population.
In other words, it evaluates how close the given sample comes to meeting the attack requirements.
In our attack, we evaluate each individual in the population by applying a loss function (\ie cross entropy) as the fitness function:
\begin{equation*}
    argmax_{\hat{x}} L(\hat{x}, y) \quad s.t. \quad \Delta(\hat{x}, x) \leq \epsilon
\end{equation*}
Loss values reflect the fitness scores of each sample in the population.
The desired adversarial samples have higher fitness scores, while other samples have lower fitness scores.

\textbf{\ul{Selection.}} This step helps a new generation to inherit genetic information by selecting samples.
In traditional genetic algorithms, the proportionate fitness selection technique~\cite{back1996evolutionary} is widely used, in which samples in the population with high fitness scores have higher chances of propagating their features to the next generation.
The technique helps generate better adversarial samples to fool a DNN model, but those samples can have higher chances of being detectable when an interpreter is applied.
In our case, our main objective is to generate adversarial samples that are undetectable by an interpreter.
Based on the objective, we randomly select two samples from the population, one of which has a higher fitness score, and the second has a lower fitness score.
We refer to them as ``winner'' (larger fitness scores) and ``loser''.
By doing that, we try to keep the perturbation in the newly generated sample's area that is considered important by the target DNN model and its interpreter.
After the process, the selected samples are passed to the crossover phase.

\textbf{\ul{Crossover.}} The phase is also called recombination.
The function is used to combine the genetic information from the selected parents to produce new offspring.
In traditional genetic algorithms, two individuals with high fitness scores are selected to pass on their genetic information to a new offspring.
In our case, we have a winner and a loser after the selection process.
We generate a new offspring (adversarial sample) by transferring the genetic data of the winner and the loser with the predefined crossover rate $cr$: $\psi_{child} = \psi_{winner} \ast S_{cr} + \psi_{loser} \ast (1 - S_{cr})$, where $S_{cr}$ is a matrix with values of 1 and 0. $\psi_{winner}$ and $\psi_{loser}$ are the samples that we selected in the previous step. $S_{cr}$ is generated based on the crossover rate $cr$ as follows:
\begin{equation*}
\begin{split}
    S_{cr} = \left\{ \begin{array}{lcr}
         1 \quad rand(0, 1) < cr \\
         0 \quad otherwise
         \end{array}\right.,
\end{split}
\end{equation*}
where $rand(0, 1)$ generates uniformly distributed numbers between 0 and 1, $cr$ is the probability of crossover (\ie the crossover rate).
In the experiment, we use the default value for MGA's crossover rate, which is 0.7.

\begin{figure}[ht]
    \centering
    \includegraphics[width=0.7\linewidth]{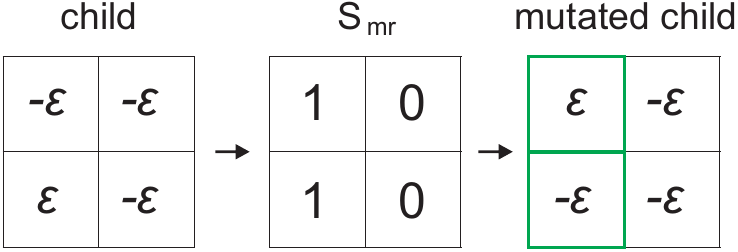}
    \caption{Example illustration of the mutation process. $S_{mr}$ is the binary matrix that is generated based on mutation probability.}
    \label{fig:mutation}
\end{figure}

\textbf{\ul{Mutation.}} The process diversifies the population and helps reach the points outside the regions of the local optima~\cite{corus2017easiest}.
The attack requires that there is enough diversity in the population, which can be introduced by the mutation function.
Mutation can be carried out via binary encoding: $\psi_{child} = -\psi_{child} \ast S_{mr} + \psi_{child} \ast (1 - S_{mr})$, where $S_{mr}$ is a binary matrix that is generated based on the mutation rate $mr$:
\begin{equation*}
\begin{split}
    S_{mr} = \left\{ \begin{array}{lcr}
         1 \quad rand(0, 1) < mr \\
         0 \quad otherwise
         \end{array}\right.,
\end{split}
\end{equation*}
where $mr$ is the probability of mutation (\ie the mutation rate).
In the experiment, we use the default value for MGA's mutation rate, which is 1e-4. \autoref{fig:mutation} illustrates the process.

\textbf{\ul{Population update.}} For continuous evolution, the population should be updated by keeping the winners and replacing the losers with new generations.
In this step, the selected loser is replaced with the mutated offspring.

The size of the population is an important parameter that directly affects the ability to search for an optimal solution in the search space.
As our attack must fulfill more requirements than traditional attacks, the adversarial search space becomes smaller.
In that case, having a large population size is not sufficient, which can lead to increased time complexity and make the search more complex by having more generations converge.
Taking this into consideration, we use 5 as the population size for our experiment.

In summary, AdvEdge generates adversarial samples against a source DNN model $f'$ and its coupled interpreter $g$ in a white-box setting.
Then we provide those generated samples as the seed for the initial population.
The algorithm evaluates the fitness scores of the population by sending them to the target DNN model $f$ (black-box).
In our case, the population size is 5.
Therefore, we waste 5 queries to evaluate.
While evaluating the initial population, if one of the individuals meets the attack requirements as the desired adversarial sample, the algorithm stops further steps.
Otherwise, it randomly chooses two individuals (a winner and a loser) from the population and calculates their fitness scores if there is any child generated from a previous iteration (this process is applied if this is not the first iteration).
After that, it generates a new offspring from the winner and the loser by conducting a crossover process.
A newly created offspring is mutated in the mutation phase.
Finally, the loser is replaced by that offspring, and the winner is kept.
The attack repeats the steps until it succeeds or reaches the query threshold.

\section{Experiments \& Results} \label{sec:evaluation}
This section provides the settings and metrics used for our experiments and the results of the proposed attack.  

\subsection{Experimental Settings and Evaluation Metrics} \label{subsec:experimental_settings}
\BfPara{Datasets} Our experiment uses ImageNet, CIFAR-10, and CIFAR-100 datasets, which cover 1.2 million images for 1,000 categories, 60,000 images for 10 categories and 60,000 images for 100 categories, respectively.
We randomly select an image from each category of the validation set in ImageNet and testing sets in CIFAR-10 and 100 (overall 3,000 images, 1,000 images per dataset), which are correctly classified by the selected DNN model $f$ with a classification confidence score higher than 60\%.
In the experiment, we set $\epsilon$ to 8 for the datasets in the scale of [0, 225], which is similar to the settings of the AdvEdge attack~\cite{abdukhamidov2021advedge}.
By experimenting with the selected images, we ensure that the proposed attack is valid across all categories of the selected DNN models.

\BfPara{Classifiers} Our experiment includes different well-known DNN models.
They include Inception-V3, DenseNet-169, VGG-19, and ResNet-50, which demonstrated high performance with 76.6\%, 77.9\%, 71.3\% and 77.2\% top-1 accuracy on ImageNet, respectively.
We use DenseNet-169 and ResNet-50 for two purposes in our evaluation. 
We adopt pretrained versions of these four models.
They are used as source DNN models (white-box models) to implement the transfer-based attack part, and they are also used as target DNN models (black-box models) to be attacked.
We note that we do not use those two models to attack themselves.
Specifically, we do not use ResNet-50 (in white-box setting) to attack ResNet-50 (in black-box setting), and the same is true for DenseNet-169 to create a realistic black-box attack scenario.
For hyperparameters, as we have a limited adversarial search space, we set the maximum query to a larger number, which is 50,000.
In the transfer-based approach, the step size $\alpha$ and the number of iterations are set to 1/255 and 300, which are the same as the AdvEdge attack settings~\cite{abdukhamidov2021advedge}.

\BfPara{Interpreters} CAM~\cite{zhou2016learning} and Grad~\cite{simonyan2014deep} interpreters are adopted as the representatives of interpretation models.
They represent different types of interpretation models, as they use different characteristics of DNN models.
CAM uses the feature maps of convolutional layers in a DNN model to generate interpretation maps: $m_{c} = \sum_{i} w_{i, c} a_{i} (j, k)$, where $a_{i}(j, k)$ is the activation of the \textit{i}th channel at the spatial location $(j, k)$ and $ w_{i, c}$ is the weight of the \textit{i}th input and the \textit{c}th output in the linear layer of a DNN model.
Grad calculates the gradients of a prediction of a DNN model based on a sample input: $m = \left| \frac{\partial f_{y} (x)}{\partial x} \right|$.
We set $\lambda$ in \autoref{eq:mainFormula} at 0.007 and 0.204 for Grad and CAM, respectively~\cite{abdukhamidov2021advedge}.
We use their open-source implementations for the experiment.





\BfPara{Evaluation Metrics} We apply different evaluation metrics as we use different DNN classifiers and interpreters.
The following metrics are used to evaluate the attack.

\begin{itemize}[leftmargin=1em]
    \item \textbf{Attack success rate:} Calculates the ratio of successful attack cases to total attack cases: $\{\# \text{successful\_test\_cases}\} \div  \{\# \text{total\_test\_cases}\}$.
    A successful test case is one that causes the target model to misclassify an image, which is the ultimate goal of the attack.
    A higher success rate indicates that the attack algorithm is more effective in generating successful adversarial examples.

    \item \textbf{Average queries:} The efficiency of the attack algorithm is evaluated using the metric of the average number of queries required to generate successful adversarial examples in a black-box setting.
    This metric is important as it reflects the amount of information the attacker needs to obtain from the target model.
    A lower average number of queries indicates that the attack algorithm is more efficient in generating adversarial examples with limited access to the target model. 
    \item \textbf{Noise rate:} The noise rate metric is used to evaluate the quality of adversarial examples generated by the attack algorithm.
    It measures the amount of noise that needs to be added to the original image to create a successful adversarial example.
    A lower noise rate indicates that the attack algorithm is more effective in generating high-quality adversarial examples with smaller amounts of noise.
    The amount of disturbance is calculated using the structural similarity index (SSIM)~\cite{wang2004image}.
    SSIM measures the similarity score, and we find the nonsimilarity portion using that score (\ie $\text{noise\_rate} = 1 - \text{SSIM}$).
\end{itemize}

To evaluate the effectiveness of the attack against interpreters, we employ the following metrics.

\begin{itemize}[leftmargin=1em]
    \item \textbf{Qualitative comparison:} The metric 
    measures the ability of the attack algorithm to generate adversarial examples that are difficult to distinguish from benign cases based on the visual inspection of images and their interpretations.
    
    \item \textbf{IoU Test:} (\textbf{I}ntersection-\textbf{o}ver-\textbf{U}nion) We use the metric to measure the similarity of interpretation maps.
    In the metric, interpretation maps are converted into binary-valued maps based on a threshold to compare adversarial maps with benign ones.
    We use different numbers as threshold values (from 0.1 to 1.0), resulting in an overall 9 threshold values for each interpretation map.
    Then we measure the IoU scores using those threshold values per interpretation map and calculate their average.
\end{itemize}


We use these metrics to calculate the effectiveness of the attack against DNN classifiers with and without defenses.

\begin{table}
\caption{Success rate, average queries, and average noise of the proposed black-box attack against different classifiers and interpreters testing on 1,000 images for each dataset (total 3,000 images). The attacks provided a median query of 5.}
\centering
\label{tab:ASR}
\resizebox{0.95\linewidth}{!}{%
\begin{tabular}{c|c|c|c|c|c} 
\toprule
\rowcolor[rgb]{0.718,0.718,0.718}  
\textbf{Interpreter} & 
\multirow{-1.5}{*}{{\cellcolor[rgb]{0.718,0.718,0.718}}\makecell{\textbf{Source}\\ \textbf{Model}}}   & 
\multirow{-1.5}{*}{{\cellcolor[rgb]{0.718,0.718,0.718}} \makecell{\textbf{Target}\\\textbf{Model}}} & \makecell{\textbf{Success}\\\textbf{Rate}} & \makecell{\textbf{Avg.}\\\textbf{Queries}} &  \makecell{\textbf{Avg. Noise}\\\textbf{Rate}}  \\ 
\midrule
\rowcolor{gray!30}\multicolumn{6}{c}{\textbf{ImageNet}}                                                                                                                                                                                   \\ 
\midrule
\multirow{6}{*}{\textbf{CAM}}                          & \multirow{3}{*}{\textbf{ResNet}}   & InceptionV3           & 0.95                  & 438.24                                   & 0.21 $\pm$ 0.06               \\ 

                                                       &                                    & DenseNet              & 0.99                  & 209.76                                    & 0.20 $\pm$ 0.06               \\ 

                                                       &                                    & VGG                   & 1.00                  & 179.80                                    & 0.20 $\pm$ 0.06               \\ 
                                                       & \multirow{3}{*}{\textbf{DenseNet}} & InceptionV3           & 0.95                  & 363.31                                   & 0.21 $\pm$ 0.06               \\ 

                                                       &                                    & ResNet                & 1.00                  & 188.53                                   & 0.20 $\pm$ 0.06               \\ 

                                                       &                                    & VGG                   & 1.00                  & 158.33                                   & 0.20 $\pm$ 0.06               \\ 
\midrule
\multirow{6}{*}{\textbf{GS}}                           & \multirow{3}{*}{\textbf{ResNet}}   & InceptionV3           & 0.95                  & 479.93                                   & 0.21 $\pm$ 0.06               \\ 

                                                       &                                    & DenseNet              & 1.00                  & 231.62                                    & 0.21 $\pm$ 0.06               \\ 

                                                       &                                    & VGG                   & 1.00                  & 180.04                                    & 0.20 $\pm$ 0.06               \\ 

                                                       & \multirow{3}{*}{\textbf{DenseNet}} & InceptionV3           & 0.95                  & 372.12                                    & 0.21 $\pm$ 0.06               \\ 

                                                       &                                    & ResNet                & 1.00                  & 189.08                                    & 0.20 $\pm$ 0.06               \\ 

                                                       &                                    & VGG                   & 1.00                  & 161.25                                   & 0.20 $\pm$ 0.06               \\ 
\midrule
\rowcolor{gray!30}\multicolumn{6}{c}{\textbf{CIFAR-100}}                                                                                                                                                                                  \\ 
\midrule
\multirow{6}{*}{\textbf{CAM}}                          & \multirow{3}{*}{\textbf{ResNet}}   & InceptionV3           & 1.00                  & 145.00                                   & 0.04 $\pm$ 0.03               \\ 

                                                       &                                    & DenseNet              & 1.00                  & 70.36                                     & 0.04 $\pm$ 0.03               \\ 

                                                       &                                    & VGG                   & 1.00                  & 60.31                                    & 0.04 $\pm$ 0.03               \\ 

                                                       & \multirow{3}{*}{\textbf{DenseNet}} & InceptionV3           & 1.00                  & 120.87                                   & 0.04 $\pm$ 0.03               \\ 

                                                       &                                    & ResNet                & 1.00                  & 61.24                                     & 0.04 $\pm$ 0.03               \\ 

                                                       &                                    & VGG                   & 1.00                  & 53.11                                     & 0.04 $\pm$ 0.03               \\ 
\midrule
\multirow{6}{*}{\textbf{GS}}                           & \multirow{3}{*}{\textbf{ResNet}}   & InceptionV3           & 1.00                  & 154.98                                 & 0.04 $\pm$ 0.03               \\ 

                                                       &                                    & DenseNet              & 1.00                  & 77.69                                     & 0.04 $\pm$ 0.03               \\ 

                                                       &                                    & VGG                   & 1.00                  & 60.39                                     & 0.04 $\pm$ 0.03               \\ 

                                                       & \multirow{3}{*}{\textbf{DenseNet}} & InceptionV3           & 1.00                  & 122.82                                    & 0.04 $\pm$ 0.03               \\ 

                                                       &                                    & ResNet                & 1.00                  & 65.42                                     & 0.04 $\pm$ 0.03               \\ 

                                                       &                                    & VGG                   & 1.00                  & 56.09                                    & 0.04 $\pm$ 0.03               \\ 
\midrule
\rowcolor{gray!30}\multicolumn{6}{c}{\textbf{CIFAR-10}}                                                                                                                                                                                   \\ 
\midrule
\multirow{6}{*}{\textbf{CAM}}                          & \multirow{3}{*}{\textbf{ResNet}}   & InceptionV3           & 1.00                  & 153.03                                   & 0.04 $\pm$ 0.03               \\ 

                                                       &                                    & DenseNet              & 1.00                  & 78.45                                     & 0.04 $\pm$ 0.03               \\ 

                                                       &                                    & VGG                   & 1.00                  & 75.18                                   & 0.04 $\pm$ 0.03               \\ 

                                                       & \multirow{3}{*}{\textbf{DenseNet}} & InceptionV3           & 1.00                  & 138.66                                   & 0.04 $\pm$ 0.03               \\ 

                                                       &                                    & ResNet                & 1.00                  & 80.43                                     & 0.04 $\pm$ 0.03               \\ 

                                                       &                                    & VGG                   & 1.00                  & 67.23                                    & 0.04 $\pm$ 0.03               \\ 
\midrule
\multirow{6}{*}{\textbf{GS}}                           & \multirow{3}{*}{\textbf{ResNet}}   & InceptionV3           & 1.00                  & 169.78                                  & 0.04 $\pm$ 0.03               \\ 

                                                       &                                    & DenseNet              & 1.00                  & 86.28                                     & 0.04 $\pm$ 0.03               \\ 

                                                       &                                    & VGG                   & 1.00                  & 69.73                                    & 0.04 $\pm$ 0.03               \\ 

                                                       & \multirow{3}{*}{\textbf{DenseNet}} & InceptionV3           & 1.00                  & 140.18                                     & 0.04 $\pm$ 0.03               \\ 

                                                       &                                    & ResNet                & 1.00                  & 76.82                                     & 0.04 $\pm$ 0.03               \\ 

                                                       &                                    & VGG                   & 1.00                  &  72.16                                & 0.04 $\pm$ 0.03               \\
\bottomrule
\end{tabular}
}
\end{table}

\begin{figure*}[ht]
    \centering
    \includegraphics[width=0.8\linewidth]{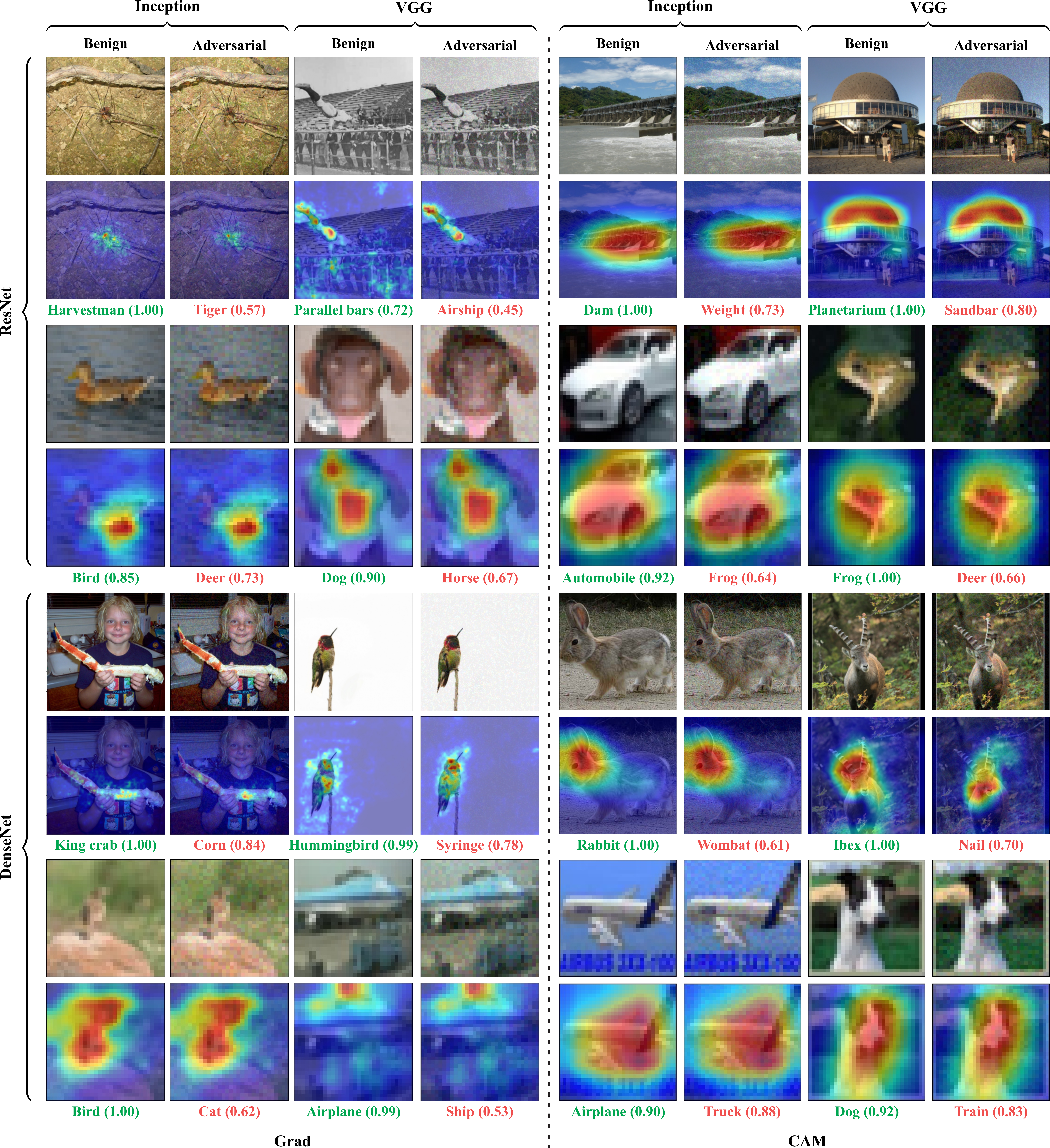}
    \caption{Attribution maps of benign and adversarial samples using \ours{} with respect to Grad and CAM on ResNet, DenseNet as source models, and Inception, VGG as target models. Samples were randomly selected from ImageNet and CIFAR datasets.}
    \label{fig:blackbox_advedge_example}\vspace{-2ex}
\end{figure*}

\subsection{Attack Effectiveness against DNNs}
\label{subsec:attack_dnns}
In this section, we discuss the effectiveness of our proposed attack on a set of four popular DNNs as target models, namely Inception-V3, ResNet, DenseNet, and VGG. Our attack is implemented and tested on two interpreters with two source models as follows: (1) CAM with ResNet, (2) CAM with DenseNet, (3) Grad with ResNet, and (4) Grad with DenseNet.  \autoref{tab:ASR} reports the results of our experiments on the scenarios mentioned above.   

\BfPara{CAM interpreter with ResNet} For the ImageNet dataset, the attack achieved a success rate of 0.95, 0.99, and 1.00 for the InceptionV3, DenseNet, and VGG target models, respectively, with an average number of queries required to execute the attack ranging from 179.80 to 438.24 queries.
The average noise rate for the attack was 0.20 $\pm 0.06$.
For the CIFAR-100 and CIFAR-10 datasets, the attack achieved a 100\% success rate for all target models, with an average number of queries ranging from 60.31 to 145.00 queries for CIFAR-100 and 75.18 to 153.03 queries for CIFAR-10.
The average noise rate for the attack was 0.04 $\pm 0.03$ for both datasets.

\BfPara{CAM interpreter with DenseNet} In terms of average queries, the CIFAR-100 dataset required the fewest queries with an average of 78.07, followed by CIFAR-10 with an average of 95.77 queries and ImageNet with the highest average of 236.72 queries.
Median queries were consistent with 5.00 for all datasets and interpreters.
Regarding the average noise rate, all datasets had similar values of 0.04 $\pm$ 0.03, except for ImageNet, which had an average noise rate of 0.20 $\pm$ 0.06.
It is noteworthy that the CIFAR datasets require fewer queries and have lower noise rates compared to the ImageNet dataset, which can be attributed to their smaller size and lower complexity.

\BfPara{Grad interpreter with ResNet} 
The attack success rate is high, with all target models achieving a success rate of 1.00 except for the ResNet-InceptionV3 combination on the ImageNet dataset which has a success rate of 0.95.
For the average queries, the CIFAR-100 dataset required the fewest queries, with an average of 97.02, followed by CIFAR-10 with an average of 105.93 queries, and ImageNet with the highest average of 297.86 queries.
Similarly, the average noise rate for the CIFAR datasets had similar values of 0.04 $\pm$ 0.03, while the ImageNet had an average noise rate of 0.20 $\pm$ 0.06.

\BfPara{Grad interpreter with DenseNet} 
For ImageNet, the attack achieved a success rate of 0.95 against InceptionV3 using the DenseNet with Grad interpreter and a success rate of 1.00 against ResNet and VGG using the same interpreter.
The average number of queries required to achieve these success rates ranged from 161.25 to 372.12, with a median of 5.00 queries.
The average noise rate of the attack was 0.20 $\pm$ 0.06.
For CIFAR-100 and CIFAR-10, the attack achieved a success rate of 1.00 against all target models using the DenseNet interpreter.
The average number of queries required ranged from 56.09 to 122.82, with a median of 5.00.
The average noise rate of the attack for both datasets was 0.04 $\pm$ 0.03.


\begin{figure}[t]
    \centering
    \includegraphics[width=1\linewidth]{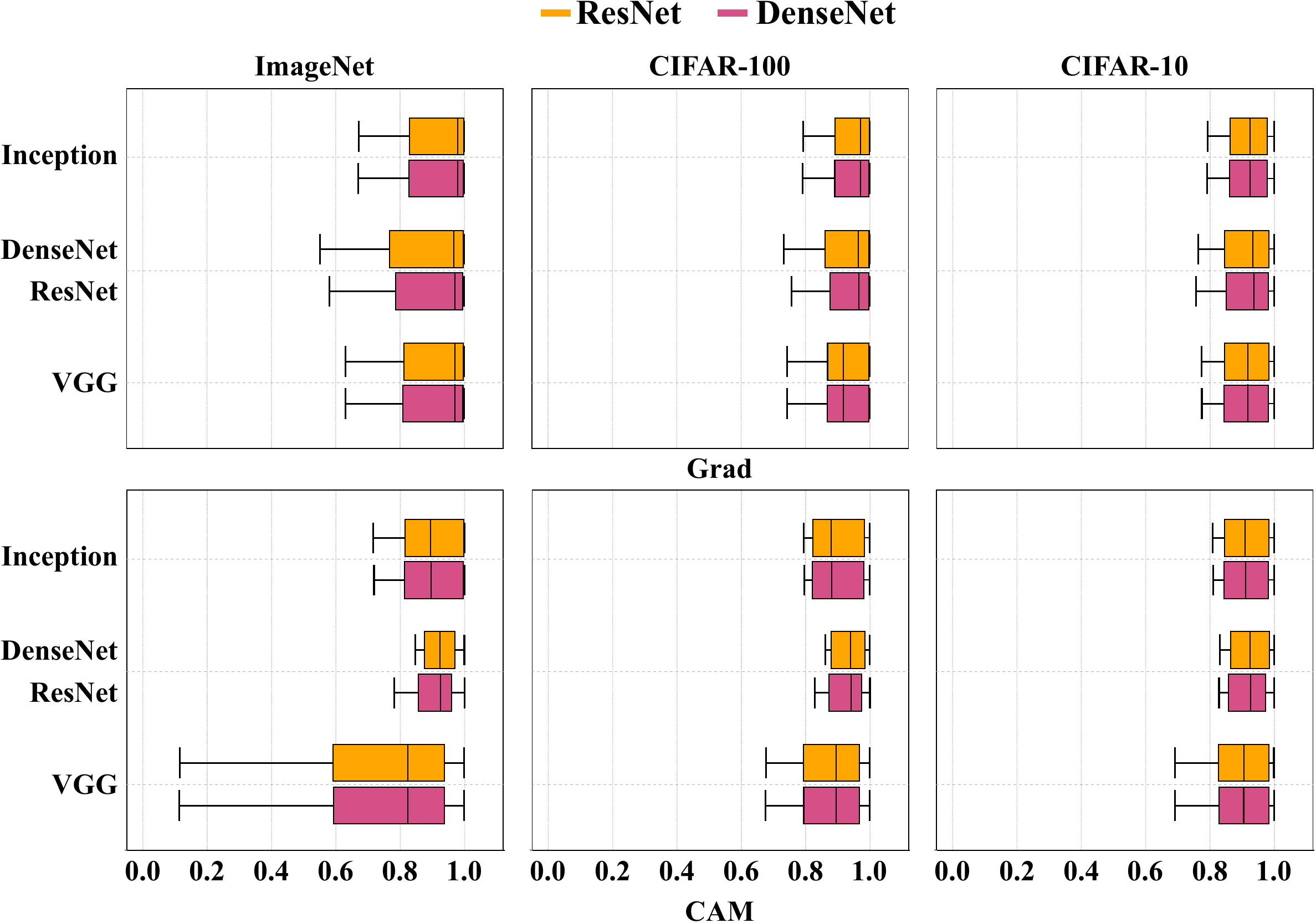}
    \caption{IoU scores of interpretation maps generated by\ours{} using Grad (top row) and CAM (bottom row) as interpreters, and ResNet and DenseNet as source models for ImageNet and CIFAR datasets. The y-axis represents the target models.}
    \label{fig:blackbox_advedge_iou}
\end{figure}

\observationdef{
\begin{observation}{Effectiveness against DNNs.}{}
The results show that the proposed attack was successful against all classifiers and interpreters, achieving a high success rate and requiring a relatively low number of queries and a low average noise rate.
  The proposed attack is effective in deceiving the DNN models with a high attack success rate using different source models.   
\end{observation}
}

\subsection{Attack Effectiveness against Interpreters} \label{subsec:attack_int}
In this part, we explore the effectiveness of our attack in comparing the benign and adversarial interpretations based on the qualitative comparison and the IoU test. 

\BfPara{Qualitative comparison} In this comparison, we differentiate whether the attribution map of our adversarial sample is similar to the attribution map of benign input.  \autoref{fig:blackbox_advedge_example} illustrates a set of examples of ImageNet and CIFAR datasets for both benign and adversarial samples with their attribution maps. Based on selected examples, it can be seen that it is difficult to distinguish between benign and adversarial attribution maps on CAM interpreter.
However, as mentioned before, the Grad interpreter is very specific in highlighting the area on the attribution map, which makes it difficult to deceive.
Even though it is challenging to mislead the Grad interpreter, our attack succeeds in generating adversarial examples that have very similar attribution maps to attribution maps of benign inputs.
Note that the attribution map of our adversarial input has less noise compared to the benign attribution map.
We can conclude that our adversarial attribution map seems more reliable than the benign one.

\BfPara{IoU Test} We evaluate the similarity between benign and adversarial attribution maps using the Intersection over Union (IoU) score.
The IoU score is calculated by finding the intersection of the maps, and as it approaches 1, the overlap between the interpretation maps becomes complete, leading to indistinguishability between them.
The performance of our attack on different DNN models with Grad and CAM interpreters is summarized in \autoref{fig:blackbox_advedge_iou}.
Our findings showed that the Grad interpreter is more effective than the CAM interpreter in conducting attacks on all datasets, with an average IoU score closer to 1. This is shown in the figure, where the median IoU score is greater than 0.9 in all datasets.
For the CAM interpreter, our attack works exceptionally well on ResNet and DenseNet, with well-balanced IoU scores and median scores above 0.9, while stability on VGG and Inception-V3 varies. 

\observationdef{
\begin{observation}{Effectiveness against Interpreters.}{}
  The proposed attack succeeds in generating adversarial interpretation maps that are indistinguishable from corresponding benign interpretation maps.   The attack shows a significant black-box attack capability with median IoU scores greater than 0.8. 
\end{observation}
}

\begin{figure*}[t]
    \centering
    \includegraphics[width=0.7\linewidth]{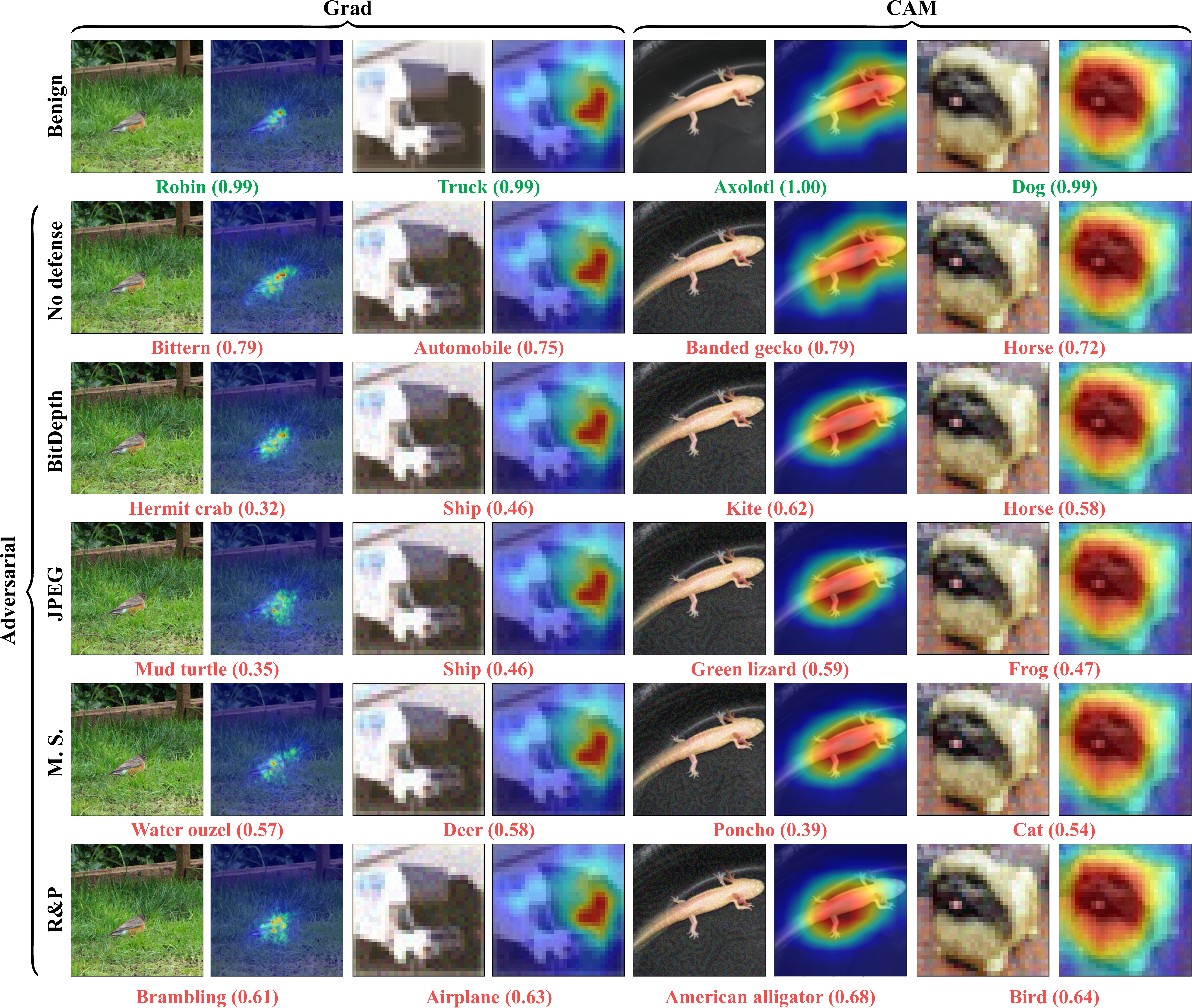}
    \caption{Adversarial samples with attribution maps generated the proposed attack with/without defense techniques using Grad, CAM on ResNet, DenseNet as source, Inception, and VGG as target models. The examples were selected at random from the ImageNet and CIFAR datasets. In the figure, M.S. stands for Median Smoothing.}
    \label{fig:defense_comparison}
    \vspace{-2ex}
\end{figure*}

\begin{figure*}[ht]
    \centering
    \includegraphics[width=0.8\linewidth]{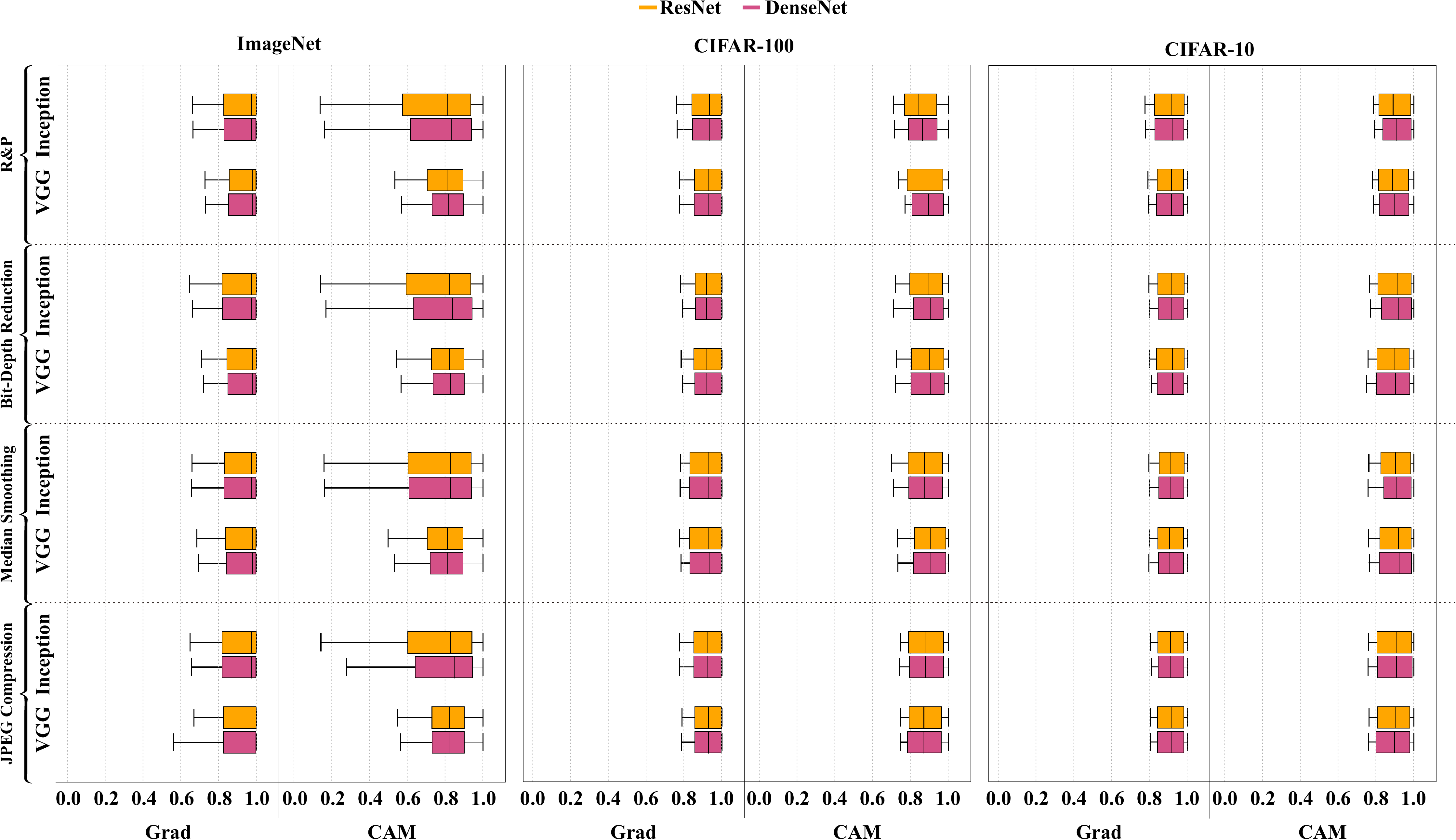}
    \caption{IoU scores of interpretation maps generated by \ours{} against four different defenses 
     using Grad and CAM interpreters with Inception-V3 and VGG as target models and ResNet and DenseNet as source models on ImageNet and CIFAR datasets.
    }
    \label{fig:blackbox_advedge_iou_defense}
\end{figure*}

\subsection{Attack Effectiveness against Defensive IDLSes} \label{subsec:defense_classifier}
This section focuses on evaluating the performance of our attack in the presence of various defense techniques.
Specifically, we investigate the impact of four commonly employed preprocessing defense strategies, namely random resizing and padding (R\&P)~\cite{xie2017mitigating}, bit depth reduction~\cite{guo2017countering}, median smoothing~\cite{ding2019advertorch}, and JPEG compression~\cite{guo2017countering}, on the efficacy of our attack.
It is important to note that these defense techniques do not modify the underlying structure of the model; rather, they alter the input sample. \autoref{fig:defense_comparison} displays the examples of the generated adversarial samples of the attack with/without defense mechanisms.

To experiment, we selected Inception-V3 and VGG as our target models, as they possess different architectures compared to our source models (ResNet and DenseNet).
 We present the results of our attack against the aforementioned defenses in \autoref{tab:defense_results}, with all experiments performed using the default values of the defense techniques. \autoref{fig:blackbox_advedge_iou_defense} showcases the IoU scores of the interpretation maps generated by our attack against four defense techniques, using Grad and CAM interpreters, and ResNet and DenseNet as source models, InceptionV3 and VGG as target models, in the ImageNet, CIFAR-100, and CIFAR-10 datasets, respectively.

\BfPara{Random resizing and padding (R\&P)} Comparing the results of the datasets, 
we can see that in general, the success rates of the attack are lower on the ImageNet dataset compared to the CIFAR-100 and CIFAR-10 datasets.
For example, the success rates on the ImageNet dataset range from 0.82 to 0.95, while on the CIFAR datasets, the success rates range from 0.96 to 0.99.
Furthermore, we can observe that the average number of queries required to successfully execute the attack is higher on the ImageNet dataset (ranging from 77.06 to 248.9) than on the CIFAR datasets (ranging from 35.33 to 157.41). The lower success rates and the higher number of queries on the ImageNet dataset than CIFAR datasets can be attributed to a larger size, greater image diversity of the dataset and its complexity, and larger search space to generate effective adversarial perturbations.

In terms of IoU score, we observe that on the Grad interpreter, our attack performs significantly well, with consistently high IoU scores across both target models. When CAM interpreter is used, we observe that our attack's performance on Inception-V3 with ImageNet dataset is somewhat unstable, whereas the results obtained on the CIFAR dataset are more reliable. The difference can be attributed to each interpreter's interpretability and robustness characteristics.

\BfPara{Bit-Depth Reduction} 
When this defense is used, it takes more queries for the attack to produce effective adversarial samples. This suggests that the defense technique is more robust than the other defenses. However, the success rate of the attack is high, ranging from 0.96 to 1.00 and the average queries required to succeed in the attack range from 52.38 to 514.87, with the lowest average queries required for the ResNet model with the Grad interpreter against VGG and the highest average queries required for the ResNet model with the CAM interpreter against Inception-V3.
The attack with this defense showed a similar trend as R\&P in performance based on the IoU score. It performed well in the Grad interpreter, consistently achieving high scores in all target models. And the attack shows instability in the CAM interpreter with Inception-V3 and the ImageNet dataset.

\BfPara{Median Smoothing} Regarding the success rate, the attack has a high success rate for all datasets ranging from 0.96 to 1.00.
Regarding the average queries, the attack requires more queries for the ImageNet dataset than for the CIFAR-100 and CIFAR-10 datasets.
For ImageNet, the average queries range from 89.55 to 315.49, while for CIFAR-100 and CIFAR-10 datasets, the average queries range from 41.05 to 144.64 and 56.63 to 141.73, respectively. As for the median queries, the attack requires only five queries for all datasets and target models. 
Compared to previous defense methods, the technique required more queries on both datasets. This indicates that it effectively preprocesses adversarial samples.
Furthermore, when this defense is used, the attack demonstrates IoU performance comparable to previous defenses. Although it requires more queries, the attack still achieves similar IoU scores.

\BfPara{JPEG Compression} The attack success rate varies between 0.81 and 0.99, with higher success rates achieved on the CIFAR dataset than on the ImageNet dataset.
The average queries required for successful attacks ranged from 61.01 to 454.78, with the median queries being five for all combinations, indicating that the attack can successfully break the defenses with a small number of queries. 
Note that most defense techniques are designed to reduce the noise in adversarial examples. However, the average noise rate remains unchanged, around 0.21 $\pm$ 0.06 and 0.04 $\pm$ 0.03. As depicted in \autoref{fig:defense_comparison}, the added perturbation in the samples is difficult to detect visually, even with the implementation of defenses. 


\observationdef{
\begin{observation}{Effectiveness against Defensive IDLSes.}{}
  The proposed attack achieves a high success rate against IDLSes with defense techniques.
  The results suggest that the use of defenses does not have a significant impact on the performance of the attack.
\end{observation}
}

\begin{table*}[t]
    \centering
    \caption{Success rate, average queries, and average noise of proposed attack against models and interpreters when using various defenses. The results are based on 3,000 images using ImageNet and CIFAR datasets. The median of queries is 5. 
    }
    \label{tab:defense_results}
    \arrayrulecolor{black}
    \resizebox{\linewidth}{!}{%
    \begin{tabular}{c|c|c|ccc|ccc|ccc|ccc} 
    \toprule
    
    \rowcolor[rgb]{0.718,0.718,0.718} {\cellcolor[rgb]{0.718,0.718,0.718}}                                       & {\cellcolor[rgb]{0.718,0.718,0.718}}                                        & {\cellcolor[rgb]{0.718,0.718,0.718}}                                        & \multicolumn{3}{c|}{\textbf{R\&P}}                                              & \multicolumn{3}{c|}{\textbf{BitDepth}}                                          & \multicolumn{3}{c|}{\textbf{Median Smoothing}}                     & \multicolumn{3}{c}{\textbf{JPEG}}  \\
    \hhline{>{\arrayrulecolor[rgb]{0.718,0.718,0.718}}--->{\arrayrulecolor{black}}------------}

    \rowcolor[rgb]{0.718,0.718,0.718} 
    \multirow{-2}{*}{{\cellcolor[rgb]{0.718,0.718,0.718}}\textbf{Interpreter}} & 
    \multirow{-2}{*}{{\cellcolor[rgb]{0.718,0.718,0.718}} \makecell{\textbf{Source} \\ \textbf{Model}}} & 
    \multirow{-2}{*}{{\cellcolor[rgb]{0.718,0.718,0.718}} \makecell{\textbf{Target}\\ \textbf{Model}}} & 
    \makecell{\textbf{Success} \\ \textbf{Rate}} & \makecell{\textbf{Avg.}\\ \textbf{ Queries}} & \cellcolor[rgb]{0.718,0.718,0.718}\makecell{\textbf{Avg. Noise}\\ \textbf{ Rate}} & 
    \makecell{\textbf{Success}\\ \textbf{Rate}} & 
    \makecell{\textbf{Avg.}\\ \textbf{Queries}} & 
    \makecell{\textbf{Avg.Noise}\\ \textbf{Rate} }     & \makecell{\textbf{Success}\\ \textbf{Rate}} & 
    \makecell{\textbf{Avg.}\\ \textbf{Queries}} & 
    \makecell{\textbf{Avg. Noise}\\ \textbf{Rate} }& \makecell{\textbf{Success}\\ \textbf{Rate}} & 
    \makecell{\textbf{Avg.}\\ \textbf{Queries}} & 
    \makecell{\textbf{Avg. Noise}\\ \textbf{Rate}}  \\

    \midrule

    \rowcolor{gray!30}\multicolumn{15}{c}{\textbf{ImageNet}}\\  

    \midrule
    \multirow{4}{*}{\textbf{CAM}}                                                                                & \multirow{2}{*}{\textbf{ResNet}}                                            & InceptionV3                                                                 & 0.82                  & 188.49                                       & 0.22$\pm$0.05                & 0.96                  & 514.87                                    & 0.21$\pm$0.06                     & 0.96                  & 286.79                                       & 0.21$\pm$0.06                & 0.81                  & 427.31                                       & 0.22$\pm$0.06                 \\ 
                                                                                                                 &                                                                             & VGG                                                                         & 0.95                  & 78.85                                       & 0.21$\pm$0.06                & 1.00                  & 140.65                                    & 0.19$\pm$0.06                     & 1.00                  & 118.33                                     & 0.21$\pm$0.06                & 0.95                  & 139.02                                      & 0.21$\pm$0.06                 \\ 
                                                                                                                 & \multirow{2}{*}{\textbf{DenseNet}}                                          & InceptionV3                                                                 & 0.87                  & 104.45                                     & 0.21$\pm$0.06                & 0.96                  & 441.57                                     & 0.21$\pm$0.06                     & 0.97                  & 255.73                                    & 0.21$\pm$0.06                & 0.86                  & 334.05                                     & 0.21$\pm$0.06                 \\ 
                                                                                                                 &                                                                             & VGG                                                                         & 0.92                  & 77.06                                      & 0.20$\pm$0.06                & 0.99                  & 167.78                                     & 0.19$\pm$0.06                     & 1.00                  & 89.55                                      & 0.20$\pm$0.06                & 0.95                  & 163.73                                     & 0.21$\pm$0.06                 \\ 
                                                                                                                 
    \cmidrule{1-15}
    \multirow{4}{*}{\textbf{GS}}                                                                                 & \multirow{2}{*}{\textbf{ResNet}}                                            & InceptionV3                                                                 & 0.82                  & 248.9                                        & 0.22$\pm$0.06                & 0.96                  & 509.51                                      & 0.20$\pm$0.06                     & 0.97                  & 315.49                                      & 0.21$\pm$0.06                & 0.82                  & 454.78                                      & 0.22$\pm$0.06                 \\ 

                                                                                                                 &                                                                             & VGG                                                                         & 0.92                  & 152.1                                      & 0.21$\pm$0.05                & 0.99                  & 114.25                                     & 0.20$\pm$0.06                     & 1.00                  & 109.28                                     & 0.21$\pm$0.06                & 0.99                  & 133.07                                     & 0.20$\pm$0.06                 \\ 

                                                                                                                 & \multirow{2}{*}{\textbf{DenseNet}}                                          & InceptionV3                                                                 & 0.86                  & 100.53                                      & 0.21$\pm$0.06                & 0.96                  & 451.88                                    & 0.21$\pm$0.06                     & 0.97                  & 248.4                                     & 0.21$\pm$0.06                & 0.86                  & 323.8                                      & 0.21$\pm$0.06                 \\ 
 
                                                                                                                 &                                                                             & VGG                                                                         & 0.92                  & 87.83                                      & 0.20$\pm$0.06                & 0.99                  & 142.9                                      & 0.20$\pm$0.06                    & 1.00                  & 96.49                                     & 0.20$\pm$0.06                & 0.99                  & 161.93                                     & 0.20$\pm$0.06                 \\ 

    \midrule

    \rowcolor{gray!30}\multicolumn{15}{c}{\textbf{CIFAR-100}}                                                                                                                                                                                                                                                                                                                                                                                                                                            \\ 
    \midrule
    \multirow{4}{*}{\textbf{CAM}}                                                                                & \multirow{2}{*}{\textbf{ResNet}}                                            & InceptionV3                                                                 & 0.99                  & 86.41                                       & 0.04$\pm$0.03                & 0.99                  & 116.05                                      & 0.04$\pm$0.03                     & 0.98                  & 131.48                                       & 0.04$\pm$0.03                & 0.98                  & 135.90                                       & 0.04$\pm$0.03                 \\ 
 
                                                                                                                 &                                                                             & VGG                                                                         & 0.99                  & 36.15                                      & 0.04$\pm$0.03                & 0.99                  & 64.48                                      & 0.04$\pm$0.03                     & 0.98                  & 54.25                                       & 0.04$\pm$0.03                & 0.98                  & 63.73                                     & 0.04$\pm$0.03                 \\ 

                                                                                                                 & \multirow{2}{*}{\textbf{DenseNet}}                                          & InceptionV3                                                                 & 0.98                  & 47.89                                      & 0.04$\pm$0.03                & 0.99                  & 102.44                                     & 0.04$\pm$0.03                     & 0.98                  & 117.24                                    & 0.04$\pm$0.03                & 0.97                  & 113.15                                     & 0.04$\pm$0.03                 \\   
 
                                                                                                                 &                                                                             & VGG                                                                         & 0.99                  & 35.33                                      & 0.04$\pm$0.03                & 0.99                  & 76.92                                      & 0.04$\pm$0.03                     & 0.98                  & 41.05                                      & 0.04$\pm$0.03                & 0.98                  & 75.06                                     & 0.04$\pm$0.03                 \\ 
    \cmidrule{1-15}
    \multirow{4}{*}{\textbf{GS}}                                                                                 & \multirow{2}{*}{\textbf{ResNet}}                                            & InceptionV3                                                                 & 0.99                  & 114.11                                       & 0.04$\pm$0.03                & 0.99                  & 133.59                                      & 0.04$\pm$0.03                     & 0.99                  & 144.64                                      & 0.04$\pm$0.03                & 0.97                  & 108.50                                      & 0.04$\pm$0.03                 \\ 
 
                                                                                                                 &                                                                             & VGG                                                                         & 0.99                  & 69.73                                      & 0.04$\pm$0.03                & 0.99                  & 52.38                                      & 0.04$\pm$0.03                     & 0.99                  & 50.10                                      & 0.04$\pm$0.03                & 0.98                  & 61.01                                     & 0.04$\pm$0.03                 \\ 
 
                                                                                                                 & \multirow{2}{*}{\textbf{DenseNet}}                                          & InceptionV3                                                                 & 0.99                  & 46.09                                     & 0.04$\pm$0.03                & 0.99                  & 107.17                                    & 0.04$\pm$0.03                     & 0.99                  & 113.88                                     & 0.04$\pm$0.03                & 0.97                  & 128.45                                     & 0.04$\pm$0.03                 \\

                                                                                                                 &                                                                             & VGG                                                                         & 0.99                  & 40.27                                     & 0.04$\pm$0.03                & 0.99                  & 65.51                                     & 0.04$\pm$0.03                     & 0.99                  & 44.24                                     & 0.04$\pm$0.03                & 0.98                  & 74.24                                      & 0.04$\pm$0.03                 \\

    \midrule
    \rowcolor{gray!30}\multicolumn{15}{c}{\textbf{CIFAR-10}}                                                                                                                                                                                                                                                                                                                                                                                                                                           \\ 
    \midrule
    \multirow{4}{*}{\textbf{CAM}}                                                                                & \multirow{2}{*}{\textbf{ResNet}}                                            & InceptionV3                                                                 & 0.97                  & 119.21                                      & 0.04$\pm$0.03                & 0.96                  & 125.61                                       & 0.04$\pm$0.03                     & 0.97                  & 131.37                                       & 0.04$\pm$0.03                & 0.96                  & 170.24                                      & 0.04$\pm$0.03                 \\

                                                                                                                 &                                                                             & VGG                                                                         & 0.97                  & 49.87                                      & 0.04$\pm$0.03                & 0.97                  & 88.95                                     & 0.04$\pm$0.03                     & 0.97                  & 74.83                                      & 0.04$\pm$0.03                & 0.98                  & 87.92                                      & 0.04$\pm$0.03                 \\

                                                                                                                 & \multirow{2}{*}{\textbf{DenseNet}}                                          & InceptionV3                                                                 & 0.96                  & 66.06                                     & 0.04$\pm$0.03                & 0.97                  & 139.26                                    & 0.04$\pm$0.03                     & 0.96                  & 141.73                                    & 0.04$\pm$0.03                & 0.97                  & 131.26                                    & 0.04$\pm$0.03                 \\

                                                                                                                 &                                                                             & VGG                                                                         & 0.98                  & 48.73                                      & 0.04$\pm$0.03                & 0.96                  & 106.11                                     & 0.04$\pm$0.03                     & 0.97                  & 56.63                                      & 0.04$\pm$0.03                & 0.97                  & 103.55                                    & 0.04$\pm$0.03                 \\ 
 
    \cmidrule{1-15}
    \multirow{4}{*}{\textbf{GS}}                                                                                 & \multirow{2}{*}{\textbf{ResNet}}                                            & InceptionV3                                                                 & 0.98                  & 157.41                                       & 0.04$\pm$0.03                & 0.97                  & 122.22                                       & 0.04$\pm$0.03                     & 0.96                  & 139.52                                       & 0.04$\pm$0.03                & 0.97                  & 187.61                                      & 0.04$\pm$0.03                 \\

                                                                                                                 &                                                                             & VGG                                                                         & 0.98                  & 96.19                                      & 0.04$\pm$0.03                & 0.96                  & 72.25                                     & 0.04$\pm$0.03                     & 0.97                  & 69.11                                     & 0.04$\pm$0.03                & 0.97                  & 84.16                                     & 0.04$\pm$0.03                 \\

                                                                                                                 & \multirow{2}{*}{\textbf{DenseNet}}                                          & InceptionV3                                                                 & 0.97                  & 63.58                                     & 0.04$\pm$0.03                & 0.97                  & 125.78                                     & 0.04$\pm$0.03                     & 0.97                  & 137.09                                     & 0.04$\pm$0.03                & 0.96                  & 144.78                                    & 0.04$\pm$0.03                 \\

                                                                                                                 &                                                                             & VGG                                                                         & 0.98                  & 55.55                                      & 0.04$\pm$0.03                & 0.96                  & 90.37                                      & 0.04$\pm$0.03                     & 0.98                  & 61.02                                        & 0.04$\pm$0.03                & 0.97                  & 102.41                                     & 0.04$\pm$0.03                 \\

    \bottomrule
    \end{tabular}
    }
    \end{table*}

\begin{figure*}
    \centering
    \includegraphics[width=0.8\linewidth]{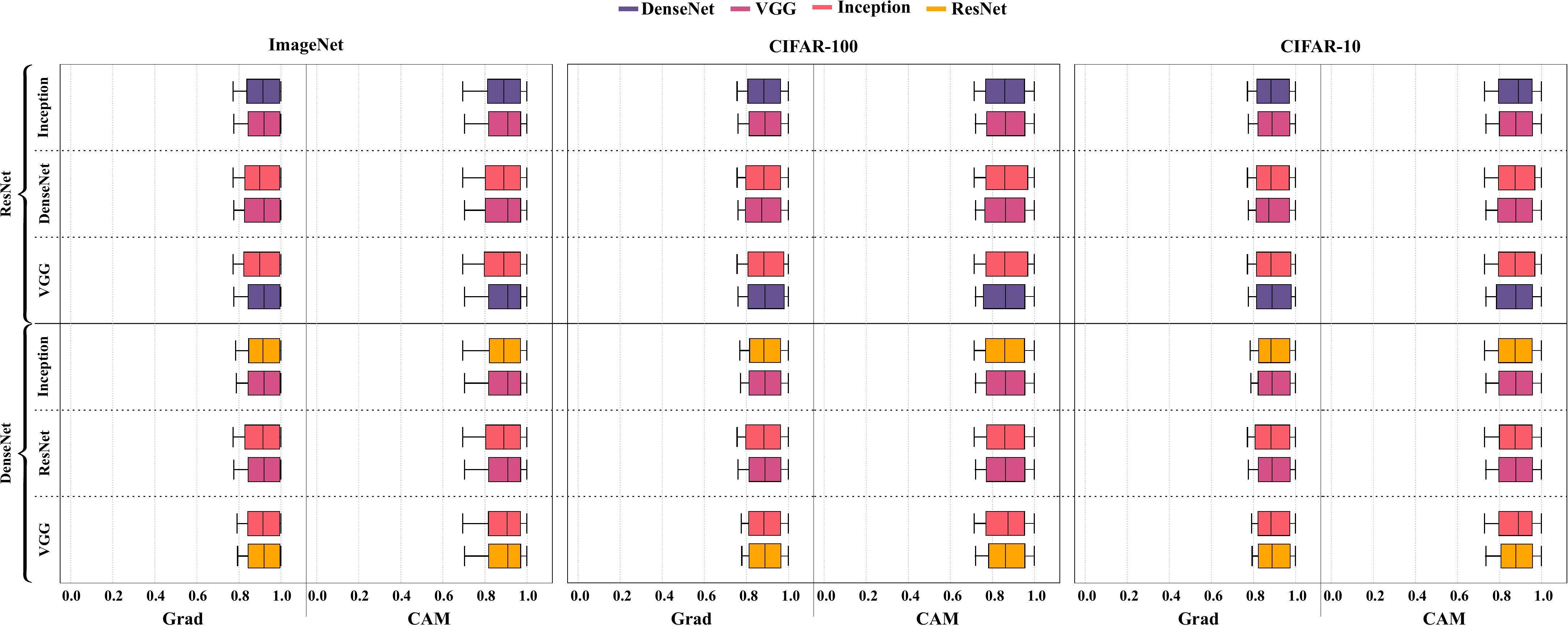}
    \caption{IoU scores of interpretation maps generated by the transferred models. The adversarial samples, originally intended to compromise given models, are transferred to alternative models using the ImageNet and CIFAR datasets.
    }
    \label{fig:transferability_datasets}
\end{figure*}


\begin{table}[t]
\caption{Success rate of adversarial samples in terms of attack transferability. Adversarial samples generated against specific DNN models are transferred to alternative DNN models on ImageNet, CIFAR-100, and CIFAR-10 datasets.}
\label{tab:transferability}
\centering
\arrayrulecolor{black}
\resizebox{\linewidth}{!}{%
\begin{tabular}{cc|c|c|ccc} 
\toprule
\rowcolor[rgb]{0.718,0.718,0.718} {\cellcolor[rgb]{0.718,0.718,0.718}}                                       & {\cellcolor[rgb]{0.718,0.718,0.718}}                                                                                                                                & {\cellcolor[rgb]{0.718,0.718,0.718}}                                                                                                                                & {\cellcolor[rgb]{0.718,0.718,0.718}}                                                                                                                                     & \multicolumn{3}{c|}{\textbf{Success Rate}}                  \\ 
\hhline{|>{\arrayrulecolor[rgb]{0.718,0.718,0.718}}---->{\arrayrulecolor{black}}---|}
\rowcolor[rgb]{0.718,0.718,0.718} \multirow{-2}{*}{{\cellcolor[rgb]{0.718,0.718,0.718}}\textbf{Interpreter}} & \multirow{-2}{*}{{\cellcolor[rgb]{0.718,0.718,0.718}}\begin{tabular}[c]{@{}>{\cellcolor[rgb]{0.718,0.718,0.718}}c@{}}\textbf{Source }\\\textbf{Model}\end{tabular}} & \multirow{-2}{*}{{\cellcolor[rgb]{0.718,0.718,0.718}}\begin{tabular}[c]{@{}>{\cellcolor[rgb]{0.718,0.718,0.718}}c@{}}\textbf{Target }\\\textbf{Model}\end{tabular}} & \multirow{-2}{*}{{\cellcolor[rgb]{0.718,0.718,0.718}}\begin{tabular}[c]{@{}>{\cellcolor[rgb]{0.718,0.718,0.718}}c@{}}\textbf{Transferred }\\\textbf{Model}\end{tabular}} & \textbf{ImageNet} & \textbf{CIFAR-100} & \textbf{CIFAR-10}  \\ 
\midrule
\multirow{12}{*}{\textbf{CAM}}                                                                               & \multirow{6}{*}{\textbf{ResNet}}                                                                                                                                    & \multirow{2}{*}{InceptionV3}                                                                                                                                        & DenseNet                                                                                                                                                                 & 0.75              & 0.72               & 0.71               \\ 

                                                                                                             &                                                                                                                                                                     &                                                                                                                                                                     & VGG                                                                                                                                                                      & 0.81              & 0.85               & 0.75               \\ 
\cmidrule{3-7}
                                                                                                             &                                                                                                                                                                     & \multirow{2}{*}{DenseNet}                                                                                                                                           & InceptionV3                                                                                                                                                              & 0.63              & 0.69               & 0.62               \\ 

                                                                                                             &                                                                                                                                                                     &                                                                                                                                                                     & VGG                                                                                                                                                                      & 0.76              & 0.71               & 0.66               \\ 
\cmidrule{3-7}
                                                                                                             &                                                                                                                                                                     & \multirow{2}{*}{VGG}                                                                                                                                                & InceptionV3                                                                                                                                                              & 0.65              & 0.64               & 0.61               \\ 

                                                                                                             &                                                                                                                                                                     &                                                                                                                                                                     & DenseNet                                                                                                                                                                 & 0.67              & 0.66               & 0.62               \\ 
\cmidrule{2-7}
                                                                                                             & \multirow{6}{*}{\textbf{DenseNet}}                                                                                                                                  & \multirow{2}{*}{InceptionV3}                                                                                                                                        & ResNet                                                                                                                                                                   & 0.74              & 0.73               & 0.68               \\ 

                                                                                                             &                                                                                                                                                                     &                                                                                                                                                                     & VGG                                                                                                                                                                      & 0.81              & 0.85               & 0.81               \\ 
\cmidrule{3-7}
                                                                                                             &                                                                                                                                                                     & \multirow{2}{*}{ResNet}                                                                                                                                             & InceptionV3                                                                                                                                                              & 0.69              & 0.75               & 0.69               \\ 

                                                                                                             &                                                                                                                                                                     &                                                                                                                                                                     & VGG                                                                                                                                                                      & 0.71              & 0.68               & 0.59               \\ 
\cmidrule{3-7}
                                                                                                             &                                                                                                                                                                     & \multirow{2}{*}{VGG}                                                                                                                                                & InceptionV3                                                                                                                                                              & 0.62              & 0.58               & 0.54               \\ 

                                                                                                             &                                                                                                                                                                     &                                                                                                                                                                     & ResNet                                                                                                                                                                   & 0.69              & 0.63               & 0.58               \\ 
\midrule
\multirow{12}{*}{\textbf{GS}}                                                                                & \multirow{6}{*}{\textbf{ResNet}}                                                                                                                                    & \multirow{2}{*}{InceptionV3}                                                                                                                                        & DenseNet                                                                                                                                                                 & 0.83              & 0.74               & 0.64               \\ 

                                                                                                             &                                                                                                                                                                     &                                                                                                                                                                     & VGG                                                                                                                                                                      & 0.79              & 0.76               & 0.71               \\ 
\cmidrule{3-7}
                                                                                                             &                                                                                                                                                                     & \multirow{2}{*}{DenseNet}                                                                                                                                           & InceptionV3                                                                                                                                                              & 0.71              & 0.71               & 0.60               \\ 

                                                                                                             &                                                                                                                                                                     &                                                                                                                                                                     & VGG                                                                                                                                                                      & 0.73              & 0.73               & 0.64               \\ 
\cmidrule{3-7}
                                                                                                             &                                                                                                                                                                     & \multirow{2}{*}{VGG}                                                                                                                                                & InceptionV3                                                                                                                                                              & 0.65              & 0.57               & 0.56               \\ 

                                                                                                             &                                                                                                                                                                     &                                                                                                                                                                     & DenseNet                                                                                                                                                                 & 0.69              & 0.63               & 0.55               \\ 
\cmidrule{2-7}
                                                                                                             & \multirow{6}{*}{\textbf{DenseNet}}                                                                                                                                  & \multirow{2}{*}{InceptionV3}                                                                                                                                        & ResNet                                                                                                                                                                   & 0.75              & 0.79               & 0.71               \\ 

                                                                                                             &                                                                                                                                                                     &                                                                                                                                                                     & VGG                                                                                                                                                                      & 0.83              & 0.82               & 0.82               \\ 
\cmidrule{3-7}
                                                                                                             &                                                                                                                                                                     & \multirow{2}{*}{ResNet}                                                                                                                                             & InceptionV3                                                                                                                                                              & 0.65              & 0.63               & 0.48               \\ 

                                                                                                             &                                                                                                                                                                     &                                                                                                                                                                     & VGG                                                                                                                                                                      & 0.72              & 0.69               & 0.63               \\ 
\cmidrule{3-7}
                                                                                                             &                                                                                                                                                                     & \multirow{2}{*}{VGG}                                                                                                                                                & InceptionV3                                                                                                                                                              & 0.62              & 0.64               & 0.57               \\ 

                                                                                                             &                                                                                                                                                                     &                                                                                                                                                                     & ResNet                                                                                                                                                                   & 0.65              & 0.69               & 0.59               \\
\bottomrule
\end{tabular}
}
\end{table}

\subsection{Attack Transferability}\label{subsec:attack-transferability}

\autoref{tab:transferability} provides the success rate of adversarial samples generated against specific DNN models as target models using source models, which are then transferred to alternative DNN models on ImageNet, CIFAR-100, and CIFAR-10 datasets.
 The results showed that the proposed attack method exhibits high transferability across different datasets with a high success rate.
 Specifically, the attack success rate in transferability achieved over 50\% across all datasets, indicating the effectiveness of the proposed method in generating adversarial examples that can fool different models trained on different datasets.  \autoref{fig:transferability_datasets} shows the IoU scores of the interpretation maps of adversarial samples regarding transferability.
 The scores indicate the similarity between the interpretation maps of adversarial samples and their benign ones.
 The results indicate that the attack transferability across all datasets and models is high.
 Specifically, the IoU scores of the interpretation maps of adversarial samples are similar to those of their benign counterparts.
 This indicates that adversarial samples are highly transferable across DNN models.

\observationdef{ 
\begin{observation}{Attack Transferability.}{}
  The proposed attack shows high transferability across various DNN models while maintaining high similarity in interpretation maps using different datasets.
\end{observation}
}

\begin{table}
\caption{Average classification confidence scores of the proposed attack against different classifiers and interpreters testing on 1,000 images for each dataset (total 3,000 images).}
\label{tab:confidence}
\centering
\arrayrulecolor{black}
\resizebox{\linewidth}{!}{%
\begin{tabular}{cccccccc} 
\toprule
\rowcolor[rgb]{0.753,0.753,0.753} {\cellcolor[rgb]{0.753,0.753,0.753}}                                          & {\cellcolor[rgb]{0.753,0.753,0.753}}                                          & \multicolumn{2}{c}{\textbf{ImageNet}} & \multicolumn{2}{c}{\textbf{CIFAR-100}} & \multicolumn{2}{c}{\textbf{CIFAR-10}}  \\ 
\hhline{|>{\arrayrulecolor[rgb]{0.753,0.753,0.753}}-->{\arrayrulecolor{black}}------}
\rowcolor[rgb]{0.753,0.753,0.753} \multirow{-2}{*}{{\cellcolor[rgb]{0.753,0.753,0.753}}\textbf{~ Source Model}} & \multirow{-2}{*}{{\cellcolor[rgb]{0.753,0.753,0.753}}\textbf{~ Target Model}} & \textbf{CAM} & \textbf{Grad}             & \textbf{CAM} & \textbf{Grad}              & \textbf{CAM} & \textbf{Grad}              \\ 
\midrule
\multirow{3}{*}{\textbf{ResNet}}                                                                                & InceptionV3                                                                   & 0.46         & 0.45                    & 0.53         & 0.51                     & 0.55         & 0.51                     \\ 
\cmidrule{2-8}
                                                                                                                & DenseNet                                                                      & 0.52         & 0.49                    & 0.56         & 0.54                     & 0.58         & 0.56                     \\ 
\cmidrule{2-8}
                                                                                                                & VGG                                                                           & 0.41         & 0.42                    & 0.49         & 0.50                     & 0.51         & 0.52                     \\ 
\midrule
\multirow{3}{*}{\textbf{DenseNet}}                                                                              & InceptionV3                                                                   & 0.42         & 0.43                    & 0.51         & 0.52                     & 0.53         & 0.54                     \\ 
\cmidrule{2-8}
                                                                                                                & ResNet                                                                        & 0.55         & 0.54                    & 0.57         & 0.55                     & 0.58         & 0.57                     \\ 
\cmidrule{2-8}
                                                                                                                & VGG                                                                           & 0.44         & 0.43                    & 0.48         & 0.49                     & 0.50         & 0.51                     \\
\bottomrule
\end{tabular}
}
\end{table}

\section{Discussion} \label{sec:discussion}

Although our evaluation in \autoref{sec:evaluation} demonstrates the effectiveness of \ours across a range of DNN classifiers and interpretation models, it is important to acknowledge its limitations and potential countermeasures.
We discuss these factors in the following sections.

\subsection{Limitations}\label{subsec:limitations}

During the initial development of \ours, the attack was designed to target DNN models solely on the basis of the success rate, disregarding the confidence level of the model's misclassification.
While the attack achieved a high success rate in fooling the target models (as shown in~\autoref{tab:ASR}), this is done with low misclassification confidence, as shown in~\autoref{tab:confidence}. 
This is common in black-box settings, as the objective function focuses on shifting the decision of adversarial samples to other classes, which naturally results in low misclassification confidence scores. 
The issue can be addressed by changing the objective function to target higher misclassification confidence. This means optimizing the attack to maximize the confidence of the adversarial class, which can lead to higher misclassification confidence scores.  


Another limitation of \ours is that the number of queries required to perform the attack increases with the complexity of the target DNN model. As a result, there are cases where \ours fails to fool more complex architectures. 
For example, Inception-V3 requires more queries and has a lower success rate than the other models. One possible solution to address the limitation can be optimizing the transfer-based aspect of the attack and adjusting key hyperparameters such as the perturbation threshold and maximum query count. In addition, the use of more complex source models may help mitigate the limitation. 
Although our attack requires slightly more queries than existing approaches~\cite{wang2020mgaattack, alzantot2019genattack} (\eg 98 and 130 queries for VGG and Inception-V3, respectively), the difference is not significant considering the constraints on the adversarial search space of our attack.

\subsection{Potential Countermeasures and Future Work} \label{subsec:potential-countermeasures}

Based on the illustration in \autoref{fig:blackbox_advedge_iou}, adversarial samples generated by \ours provide high-quality interpretations that are indistinguishable from their benign interpretations.
Another countermeasure is to train a DNN model with an adversarial dataset to increase its robustness. It becomes more difficult to fool a robust DNN model in a black-box setting when adversarial training is used \cite{NEURIPS2019_7503cfac, hsiung2023towards}.
Even if an attacker succeeds in fooling the adversarially trained DNN model, the perturbation added to the sample could reveal the attacker's presence. Adopting several defense techniques \cite{9049702} at the same time as an ensemble-based method ({\ie} preprocessing techniques) for the decision-making process could be another countermeasure against the attack.
The main reason for this is that defense techniques utilize various features of a sample to remove added perturbation.
Also, optimally generating an adversarial sample considering all the features of a sample becomes computationally expensive.


\section{Conclusion} \label{sec:conc}
In this work, we propose the black-box version of the AdvEdge attack, which can deceive both DNN models and their interpreters in a black-box setting.
Our attack is based on transfer-based and score-based methods and is both gradient-free and query-efficient.
Our experimental results demonstrate that our method achieves a high attack success rate and generates adversarial interpretation maps that are highly similar to benign interpretations.
Moreover, our attack is effective against various defense techniques, even when they are involved in the process.
We achieve a high attack success rate and transferability while still misleading the target interpreters, highlighting the efficacy and robustness of our proposed attack.
These findings emphasize the need to develop more robust defense mechanisms to enhance the security of DNN models against adversarial attacks.


\balance
\bibliographystyle{IEEEtran}  
\bibliography{ref} 

\end{document}